\documentclass{sig-alternate-2013}

\newfont{\mycrnotice}{ptmr8t at 7pt}
\newfont{\myconfname}{ptmri8t at 7pt}
\let\crnotice\mycrnotice%
\let\confname\myconfname%

\permission{Permission to make digital or hard copies of all or part of this work for personal or classroom use is granted without fee provided that copies are not made or distributed for profit or commercial advantage and that copies bear this notice and the full citation on the first page. Copyrights for components of this work owned by others than the author(s) must be honored. Abstracting with credit is permitted. To copy otherwise, or republish, to post on servers or to redistribute to lists, requires prior specific permission and/or a fee. Request permissions from permissions@acm.org.}
\conferenceinfo{CIKM'14,}{November 3--7, 2014, Shanghai, China.\\
{\mycrnotice{Copyright is held by the owner/author(s). Publication rights licensed to ACM.}}}
\copyrightetc{ACM \the\acmcopyr}
\crdata{978-1-4503-2598-1/14/11\ ...\$15.00.\\
http://dx.doi.org/10.1145/2661829.2662005}

\usepackage{times}
\usepackage{graphicx} % more modern
\usepackage{subfigure} 

\usepackage{algorithm}
\usepackage{algorithmic}

\usepackage{url}
\usepackage{hyperref}

\usepackage{amssymb,amsmath}
\usepackage{enumitem}
\usepackage{multirow}

\usepackage{color}
\definecolor{positivecolour}{RGB}{215,48,39}
\definecolor{negativecolour}{RGB}{69,117,180}

\newcommand{\NAcell}{\multicolumn{2}{c|}{\scriptsize$\mbox{N/A}$}}	% this give N/A take takes 2 columns
\newcommand{\argmax}{\operatornamewithlimits{argmax}}
\begin{document}
%
% --- Author Metadata here ---
%\conferenceinfo{CIKM}{'14 Shanghai, China}
%\CopyrightYear{2007} % Allows default copyright year (20XX) to be over-ridden - IF NEED BE.
%\crdata{0-12345-67-8/90/01}  % Allows default copyright data (0-89791-88-6/97/05) to be over-ridden - IF NEED BE.
% --- End of Author Metadata ---

\title{Twitter Opinion Topic Model: \\Extracting Product Opinions from Tweets by \\Leveraging Hashtags and Sentiment Lexicon}
%\subtitle{[Extended Abstract]
%\titlenote{A full version of this paper is available as
%\textit{Author's Guide to Preparing ACM SIG Proceedings Using
%\LaTeX$2_\epsilon$\ and BibTeX} at


\numberofauthors{2} %  in this sample file, there are a *total*

\author{
\alignauthor
Kar Wai Lim\\
       \affaddr{ANU \& NICTA, Canberra, Australia}\\
%       \affaddr{Canberra, Australia}\\
       \email{karwai.lim@anu.edu.au}
\alignauthor
Wray Buntine\\
       \affaddr{Monash University, Melbourne, Australia}\\
%       \affaddr{Melbourne, Australia}\\
       \email{wray.buntine@monash.edu}
}

\maketitle
\begin{abstract}
Aspect-based opinion mining is widely applied to review data to aggregate or summarize opinions of a product, and the current state-of-the-art is achieved with Latent Dirichlet Allocation (LDA)-based model.
Although social media data like tweets are laden with opinions, their ``dirty'' nature 
(as natural language) has discouraged researchers from applying LDA-based opinion model for 
product review mining. 
Tweets are often informal, unstructured and 
lacking labeled data such as categories and ratings, making it
challenging for product opinion mining.
In this paper, we propose an LDA-based opinion model named Twitter Opinion Topic Model (TOTM) for opinion mining and sentiment analysis. 
TOTM leverages {\it hashtags}, {\it mentions}, emoticons and strong sentiment words that are present in tweets in its discovery process.
It improves opinion prediction by modeling the target-opinion interaction directly, thus discovering target specific opinion words,
neglected in existing approaches.
Moreover, we propose a new formulation of incorporating sentiment prior information into a topic 
model, by utilizing an existing public sentiment lexicon. 
This is novel in that it learns and updates with the data.
We conduct experiments on 9 million tweets on electronic products, and demonstrate the improved performance of TOTM in both quantitative evaluations and qualitative analysis. 
We show that aspect-based opinion analysis on massive volume of tweets provides useful opinions on products.
% [We conclude that tweets do indeed provide useful opinion on electronic products and we encourage more research to be done on social media using model-based approach.]
%\vspace{5pt}
%\noindent 
%(Note: supplementary material will be made available online)
\end{abstract}

%\category{I.2}{Computing Methodologies}{Artificial Intelligence}
%A category including the fourth, optional field follows...

\vspace{-2mm}

\vspace{-2mm}

\vspace{-2mm}

\section{Introduction}

When making a purchase decision, 
a key deciding factor can often be the reviews written by other consumers. 
These reviews are freely available online, however, one can rarely read all the reviews given their volume.
% and ready availability. 
This has led to various automated algorithms to mine the reviews, extracting a more digestible 
summary for a user. The task of analyzing opinions from text data such as reviews is 
known as opinion mining or opinion extraction~\cite{liu2012sentiment,INR-011}.

Among various approaches to opinion mining, {\it aspect-based opinion mining} has recently gained a lot of attention from the research community. 
Aspect-based opinion mining involves extracting the major aspects or facets from data for analysis. 
As an example, for a camera product, the aspects could be ``picture quality'', ``portability'' {\it{etc}}. 
Topic models are often used to determine the aspects through soft clustering. 
Topic models have been successfully applied to review data crawled from review websites such as Epinions.com, TripAdvisor {\it etc}.
LDA-based models are considered to be state-of-the-art for aspect-based opinion mining~\cite{Moghaddam:2012:DLM:2396761.2396863}. 

%Reviews are important in influencing a consumer's purchasing behavior, some companies have seen the opportunity to achieve higher sale figures or to promote their brand by creating fake reviews. There are even companies that help a business by crafting fraudulent reviews on popular review website for a fee. Unfortunately to the consumers, fake reviews detection is a difficult task to algorithm and even to humans\footnote{Test your ability to detect fake review here: \url{http://www.cs.uic.edu/~liub/FBS/fake-reviews.html} .}.

Besides reviews extracted from review websites, opinions from social media websites are also very 
useful, even though they are often overlooked as a source for reviews. 
Social media text is short and is regarded as ``dirty'', and hence less useful for more 
sophisticated language analysis \cite{Zhao:2011:CTT:1996889.1996934}. 
The same problem also leads to degradation when applying NLP 
tools~\cite{Ritter:2011:NER:2145432.2145595}.  
%Additionally, the text data from social media 
%websites can be hard to access, for instance, Facebook posts are not publicly accessible; Twitter's 
%tweets are only accessible through an API (which limits the rate of access) or through Firehose\footnote{\small \url{http://gnip.com/products/realtime/firehose/}} 
%(which can be expensive). 
Despite these limitations, large numbers of tweets containing opinions are generated every day and 
are very relevant for opinion mining. 
We argue that while tweets are generally unstructured, Twitter is a useful source of reviews since 
it provides a convenient platform for users to express their opinions. 
Twitter is also integrated to a person's social life, making it easier for 
users to express their opinions on products by tweeting instead of writing a review on review websites.

% Furthermore, we believe that there are also less incentives for companies to create fake reviews on social media as compared to review websites.
%  hmm, what about Obama's 500,00 fake followers ;-)

In this paper, we demonstrate the usefulness of Twitter as a source for aspect-based target-opinion mining. 
We propose a novel LDA-based opinion model that is designed for tweets, which we name Twitter Opinion Topic Model (TOTM). 
TOTM models the target-opinion interaction directly, which significantly improves opinion prediction, {\it e.g.}\ TOTM discovers {\it{`grilled'}} is positive for {\it{sausage}} but not other targets.
We note that while there are no explicit ratings and scores on tweets, tweets often contain emoticons and strong sentiment words, such as `{\it love}' and `{\it hate}'.
TOTM exploits this fact and uses the information to compensate for the lack of explicit ratings. 
Additionally, {\it hashtags} are strong indicators of topics for 
tweets~\cite{Mehrotra:2013:ILT:2484028.2484166}. 
TOTM makes use of the hashtags and {\it mentions} in tweets for tweet aggregation, which improves aspect clustering.
Modeling with TOTM also allows us to acquire additional summaries on products, which are not 
obtainable with existing models.
%(see Subsection~\ref{subsec:qualitative}).

Furthermore, we incorporate a sentiment lexicon as prior information into TOTM. 
We propose a novel formulation of how the sentiment lexicon affects the priors in TOTM. 
Our approach facilitates automatic learning of the lexicon strength based on the data; while current existing methods are {\it ad hoc} or ruled-based. 
Our formulation is shown to perform best for sentiment classification.
Additionally, we propose a different target-opinion extraction procedure that works better for tweets, discussed in Subsection~\ref{subsec:preprocessing}. 
We note that text preprocessing is important when dealing with tweets.
%Although TOTM is designed for informal text, it can also be applied to standard text.

We apply TOTM to 3 tweets corpus, showing improved performance of TOTM in model fitting and 
sentiment analysis. 
Qualitatively, we demonstrate the usefulness of TOTM in 
extracting the opinions on products from tweets. 
As large volumes of tweets laden with opinions are generated daily, real-time aspect-based opinion 
analysis allows us to obtain first-hand opinions on new products, which might not be as readily 
available from review websites.

%\highlight{[Why is this work important?]}

The rest of the paper is structured as follows. 
Section~\ref{sec:relatedWork} reviews some related work, and Section~\ref{sec:problem} provides a 
summary of our task and major contributions. 
In Section~\ref{sec:ilda}, we present Interdependent LDA (ILDA)~\cite{moghaddam2011ilda} which will 
be used as a baseline for comparison. 
We introduce TOTM in Section~\ref{sec:twilda} and the method of incorporating a lexicon in 
Section~\ref{sec:incorporatingPrior}. 
In Section~\ref{sec:modellikelihood}, we discuss TOTM's model likelihood and inference procedure, 
as well as proposing a novel hyperparameter sampling procedure. 
We then describe the data used in this paper and report on the experiments in 
Sections~\ref{sec:data} and~\ref{sec:experiment}. 
Finally, we conclude the paper in Section~\ref{sec:conclusion}.

%- but contains other form of auxiliary information that is useful
%- spam is still an issue

%\vspace{-3pt}
\section{Related Work}
\label{sec:relatedWork}

%Additional Citation:
%[ N. Gupta. Extracting descriptions of problems with product and service from twitter data. In
%Proc. of Workshop on SWSM 2011 in conjunction with SIGIR 2011, Beijing, China, July 2011.]
%[Peng Zhao, Xue Li and Ke Wang. Feature Extraction from Microblogs for Comparison of Products and Services. The 14th International Conference on Web Information System Engineering, 2013.]

%[Elaboration needed for the list of non-LDA method.]

Latent Dirichlet Allocation (LDA) is a topic model that has been extended by many for sentiment analysis. 
Notable examples based on LDA include the MaxEnt-LDA hybrid model~\cite{Zhao:2010:JMA:1870658.1870664}, Joint Sentiment Topic (JST) model~\cite{Lin:2009:JSM:1645953.1646003}, Multi-grain LDA (MG-LDA)~\cite{Titov:2008:MOR:1367497.1367513}, Interdependent LDA (ILDA)~\cite{moghaddam2011ilda}, Aspect and Sentiment Unification Model (ASUM)~\cite{Jo:2011:ASU:1935826.1935932} and Multi-Aspect Sentiment (MAS) model~\cite{Titov08ajoint}.
The Topic-Sentiment Mixture (TSM) model~\cite{Mei:2007:TSM:1242572.1242596} performs sentiment analysis by utilizing the Multinomial distribution.
These models perform aspect-based opinion analysis and they had been successfully applied to review 
data of different domains, such as electronic product, hotel and restaurant reviews. 
The task of summarizing the reviews is also known as {\it opinion aggregation}. 

To the best of our knowledge, there is no existing LDA-based opinion aggregation method that has been successfully applied to social media data such as tweets. 
Current opinion mining methods that are used on tweets tend to be {\it ad hoc} or rule-based. 
We suspect this is because tweets are generally regarded as too noisy for model-based methods to work, and also due to the fact that LDA works badly on short documents. 
Maynard et al.~\cite{maynard2012challenges} studied the challenges in developing an opinion mining 
tool for social media and they advocated the use of shallow techniques in linguistic processing of 
tweets. 
Notable non-LDA-based methods for opinion analysis include OPINE~\cite{popescu2007extracting}, which uses relaxation labeling to classify sentiment, and Opinion Digger \cite{Moghaddam:2010:ODU:1871437.1871739}, an aspect-based review miner using {\it k nearest neighbor}.
Hu and Liu~\cite{hu2004mining} performed rule-based target-opinion extraction from online product reviews, while Li et al.~\cite{Li:2010:SRM:1873781.1873855} extracted opinions from reviews using Conditional Random Fields.
On tweets, Pak and Paroubek~\cite{pak2010twitter} performed opinion analysis using a Naive Bayes 
classifier; while Liu et al.~\cite{Liu:2013:ACS:2505515.2505569} performed sentiment classification 
using an adaptive co-training SVM.
Go et al.~\cite{go2009twitter} and Davidov et al.~\cite{Davidov:2010:ESL:1944566.1944594} made use of emoticons (smileys), which were found to provide improvement for sentiment classification on tweets.
Since tweets are always short, existing work~\cite{go2009twitter, pak2010twitter, Davidov:2010:ESL:1944566.1944594, Liu:2013:ACS:2505515.2505569} tends to assume a single polarity for each tweet. 
In contrast, Jiang et al.~\cite{jiang2011target} performed target-dependent sentiment analysis, where the sentiments apply to a specific target.

Lexical information can be used to improve sentiment analysis. 
He~\cite{He:2012:ISP:2184436.2184437} used a sentiment lexicon to modify the priors of LDA for sentiment classification, though with an approach with {\it ad hoc} constants.
Li et al.~\cite{li2009non} incorporated a lexical dictionary into a non-negative matrix tri-factorization model, using a simple rule-based polarity assignment.
Refer to Ding et al.~\cite{Ding:2008:HLA:1341531.1341561} and Taboada et al.~\cite{taboada2011lexicon} for a detailed review on applying lexicon-based methods in sentiment analysis. 
Instead of a lexicon, Jagarlamudi et al.~\cite{Jagarlamudi:2012:ILP:2380816.2380844} used seeded words as lexical priors for semi-supervised topic modeling.

\vspace{-3pt}
\section{Opinion Mining Task On Tweets}
\label{sec:problem}

In this section, we describe the opinion mining problem we are tackling and outline our major contributions in solving the problem.

\vspace{-2pt}
\subsection{Problem Definition}

Given a collection of documents (tweets), our first problem is to extract {\it target-opinion} pairs from each document. 
A target-opinion pair $\langle t,o\rangle$ consists of two phrases: a {\it target} phrase $t$ which is the object being described, and an {\it opinion} phrase $o$ which is the description. 
Target phrases are usually nouns and opinion phrases are usually adjectives, examples include $\langle${\it picture quality, good}$\rangle$, $\langle${\it iPhone app, expensive}$\rangle$ {\it etc}. 
Note that a phrase can be either a collocation (multi-word phrase) or a single word. 
For simplicity, we will use `{\it word}' to mean a {\it single-word} or a {\it phrase} in this paper.

Our next problem is to group the target-opinion pairs into clusters and identify the associated sentiments. 
The produced clusters should depend on the tweet corpus, as they should represent different aspects of the corpus.
For example, given a tweet corpus which consists of various electronic products, we would like different products to be grouped into different clusters.
Each target-opinion pair is assigned 2 latent labels, the first being {\it aspect} $a$ indicating which cluster the pair belongs, the second label being {\it sentiment} $r$.
The sentiment of a target-opinion pair refers to the polarity of the opinion phrase, which can be {\it negative}, {\it neutral} or {\it positive}.

Finally, we would like to display a summary (high level view) of the obtained quadruples $\langle t, o, a, r \rangle$. 
There are many ways to do this, here we follow the standard topic modeling approach to display the top phrases. 
We inspect the target phrases given the aspects. 
We also examine the opinion phrases given the target phrases and sentiments. 
In brief, our task of opinion mining on tweets is to extract useful opinions and represent them in a format that is easy to digest. 
For example, with a tweet corpus on electronic products, we would like to discover the opinions of Twitter users on certain products, such as iPhones.

\vspace{-2pt}
\subsection{Major Contributions}

We make two major contributions as follows:
Firstly, {\it we design an LDA-based topic model} (TOTM)
for performing aspect-based target-opinion analysis on product reviews from tweets. 
TOTM is novel in that it directly models the target-opinion interaction, giving significant 
improvement in opinion prediction.
Existing aspect-based methods only model the interaction between aspects and sentiments, leaving 
the targets and opinions to be weakly associated through aspects and sentiments.
Without this explicit modeling, the existing models failed to sensibly assign opinions to targets.
For example, from a restaurant review with {\it friendly staff} and {\it delicious cake}, existing 
LDA-based opinion model failed to recognize that {\it friendly} cannot be used to describe {\it 
cake}.
Also, as mentioned in the introduction, TOTM makes use of available auxiliary variables in tweets 
(hashtags, mentions, emoticons and strong sentiment words) to improve aspect-based opinion analysis.

% (the perplexity for the opinion words are much lower, meaning the model is less surprise to see the opinion words in test set).

Secondly, {\it we propose a new formulation for incorporating a sentiment lexicon} into our topic model.
While existing methods adopt an {\it ad hoc} or ruled-based approach to incorporating sentiment 
prior, our formulation is novel in that it is learned automatically given the data. 
This is done robustly using a tuning hyperparameter that is optimized automatically.
The sentiment information is used to adjust the opinion priors in order to improve sentiment analysis. 

%For this work, we use a number of techniques consistent with the
%state-of-the-art in opinion mining.
%For instance, while we use emoticons and known strong sentiment words to improve sentiment analysis \cite{Davidov:2010:ESL:1944566.1944594, Read:2005:UER:1628960.1628969}, our approach is integrated in a topic modeling framework.   
%We also use the hashtags to improve aspect clustering \cite{Mehrotra:2013:ILT:2484028.2484166}.
%Since standard NLP tools such as dependency parsers degrade on informal text, we propose additional correction steps presented in Subsection~\ref{subsec:preprocessing}.
%with Stanford Dependency Parser \cite{de2006generating} during preprocessing. We utilize Twitter NLP \cite{owoputi2013improved} to clean up the dependency relation parsed by the aforementioned dependency parser.

\section{Baseline: Interdependent LDA}
\label{sec:ilda}

\begin{figure}[h]
%\vskip 0.2in
\begin{center}
\centerline{\includegraphics[width=0.97\columnwidth]{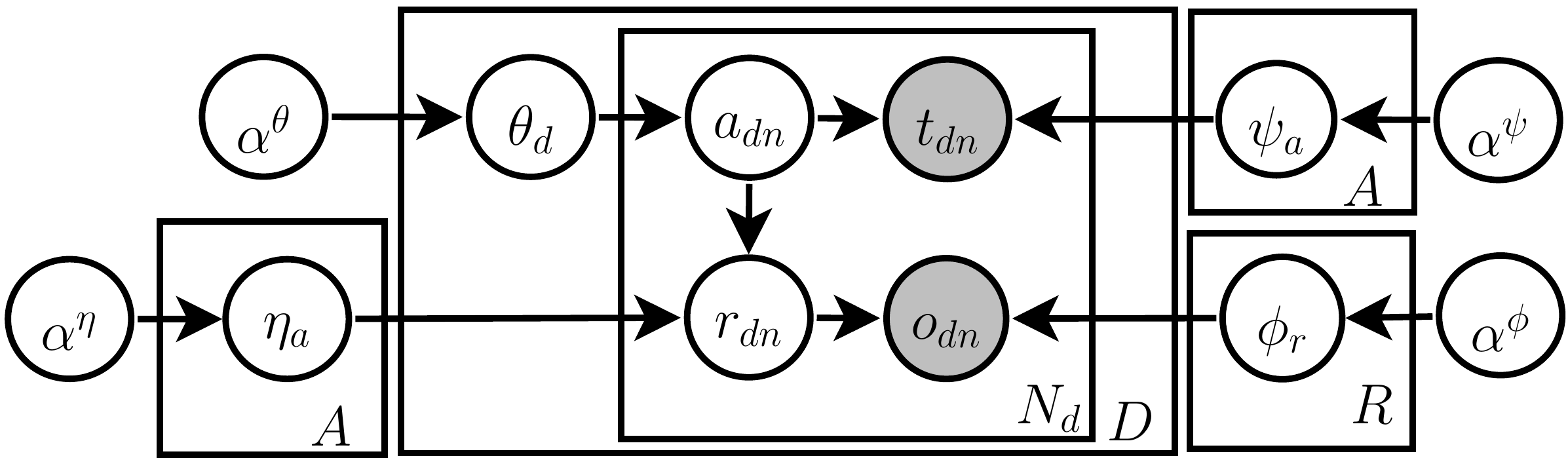}}
\caption{Graphical Model for Interdependent LDA}
\label{fig:ilda}
\end{center}
\vskip -0.275in
\end{figure}

Interdependent LDA (ILDA) \cite{moghaddam2011ilda} is an extension of LDA that performs aspect-based opinion analysis. It jointly models the aspect ($a$) and sentiment\footnote{Also known as {\it rating} in Moghaddam and Ester~\cite{moghaddam2011ilda}.} ($r$) for each target-opinion pair $\langle t,o \rangle$ that is present in a document. 
%Each target-opinion word pair consists of two words, a {\it target} word and an {\it opinion word}. 
%Target words are usually nouns and they are described by opinion words which are usually adjectives. 
%As an example, for a mobile phone product we might have $\langle$camera quality, great$\rangle$ as the target-opinion pair, with aspect {\it camera} and rating of 8, assuming that the rating takes the value from 1 to 10. 
We note that the sentiment variable $r$ is a categorical variable, and is not restricted to just 3 values. 
However, in this paper, we will assume that the sentiment $r$ has only three labels $\{-1,0,1\}$, which correspond to negative, neutral and positive sentiment respectively. 

In this paper, we treat ILDA as a baseline. 
It has the following generative process.
For each document $d$, we sample a document-aspect distribution
{
\setlength{\abovedisplayskip}{0pt}
\begin{align*}
\theta_d \sim \mbox{Dir}(\alpha^\theta)~.
\end{align*}
}For each aspect $a$, we sample an aspect-sentiment distribution $\eta_a$ and an aspect-target word distribution $\psi_a$:
\begin{align*}
\eta_a \sim \mbox{Dir}(\alpha^\eta)~,~~~~~~~~~~~~~~~~~~ 
\psi_a \sim \mbox{Dir}(\alpha^\psi)~.
\end{align*}
\noindent Given each sentiment $r$, we sample a sentiment-opinion phrase distribution 
{
\setlength{\abovedisplayskip}{0pt}
\begin{align*}
\phi_r \sim \mbox{Dir}(\alpha^\phi)~.
\end{align*}
}Finally, we model each target-opinion pair $\langle t_{dn}, o_{dn} \rangle$ and their respective latent aspects and sentiments.
\begin{align*}
a_{dn} \sim \mbox{Discrete}(\theta_d)~,~~~~~~~~~~~~~~~~~~ 
r_{dn} \sim \mbox{Discrete}(\eta_{a_{dn}})~,	\\
t_{dn} \sim \mbox{Discrete}(\psi_{a_{dn}})~,~~~~~~~~~~~~~~~~~~ 
o_{dn} \sim \mbox{Discrete}(\phi_{r_{dn}})~.
\end{align*}
\noindent
We note that the $\alpha$'s are the hyperparameters corresponding to the symmetric Dirichlet distributions.

ILDA models the sentiment conditionally on the aspect; and given the aspect and sentiment, the target word and opinion word are generated independently. 
Although such modeling is often adequate (since many of the opinion words can be applied generally 
to most target words), it fails to take into account that some opinion words are restricted to 
certain target words, and {\it vice versa}. 
For example, we can say a phone has {\it short battery life} but not {\it short camera quality}.

In this paper, we do not compare against other models such as MG-LDA and ASUM, since
these models do not perform target-based opinion analysis, and thus not directly comparable.

\section{Twitter Opinion Topic Model}
\label{sec:twilda}

\begin{table}[t!]
	\centering
	\vskip -2mm
    \caption{List of Variables for TOTM}
    \label{tbl:variables}
    \vskip 0.15in
	\begin{tabular}{|c|p{0.77\columnwidth}|}
    \hline
	\multicolumn{1}{|c|}{Variable} & \multicolumn{1}{c|}{Description} \\
	\hline
	$a$  & Aspect: category label for a target-opinion pair; also known as topic in topic models.  \\
	\hline
	$r$  & Sentiment: polarity of an opinion phrase. \\
	\hline
	$t$ & Target: word or phrase that is being described. \\
	\hline
	$o$ & Opinion: description of a target word $t$. \\
	\hline
	$e$ & Emotion Indicator: binary variable indicating positive or negative emotion; can be unobserved. \\
	\hline
	$\psi$  & Target word distribution: Probability distribution for target words. \\ 
	\hline
	$\phi, \phi', \phi^*$  & Opinion word distribution: Probability distribution for opinion words. \\
	\hline
	$\gamma$  & Sentiment distribution: Probability distribution in generating a sentiment label $r$. \\
	\hline
	$\alpha, \beta$  & Hyperparameters associated with the PYP. \\
	\hline
	$H$ & Base distribution for the PYP. \\
	\hline
	\end{tabular}
%	\vskip -0.1in
\end{table}

\begin{figure}[ht]
%\vskip -0.05in
\begin{center}
\centerline{\includegraphics[width=\columnwidth]{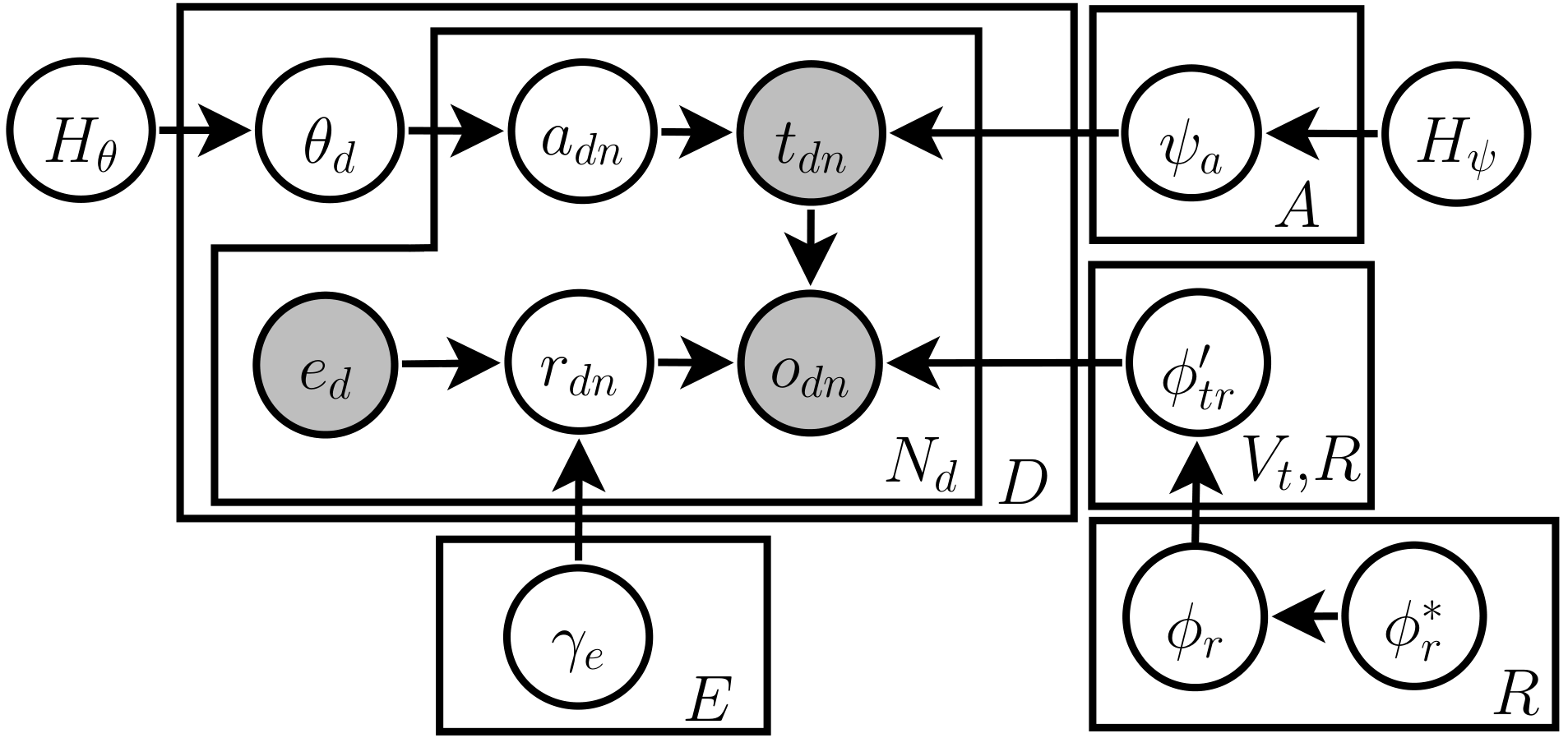}}
\caption{Graphical Model for Twitter Opinion Topic Model}
\label{fig:totm}
\end{center}
\vskip -0.275in
\end{figure} 

Here we present the Twitter Opinion Topic Model for aspect-based opinion analysis on tweets.
The model is given in Figure~\ref{fig:totm}. Contrary to ILDA, we do not model the aspect-sentiment distribution $\eta$. 
Instead, we model the target-opinion pairs directly. 
This allows us to better model the opinion words, and also provides us with a finer level of opinion analysis. 
For example, TOTM will be able to model that the word {\it{`limited'}} can describe {\it{battery life}} but is unlikely to be used to describe {\it{charger}}.

TOTM uses the {\it Griffiths-Engen-McCloskey} (GEM)~\cite{pitman1996some} distribution
to generate probability vectors and
the {\it Pitman-Yor process} (PYP)~\cite{Teh:2006:HBL:1220175.1220299}
to generate probability vector given another mean
probability vector.
Both GEM and PYP are parameterized by a discount parameter $\alpha$ and a concentration parameter $\beta$; 
and PYP is additionally parameterized by a mean or
base distribution $H$.
The GEM distribution is equivalent to the PYP with a 
base distribution that generates an ordered integer label, $H_\theta$. 
The PYP is also known as the two-parameter Poisson-Dirichlet process.

We introduce a variable $e$ named {\emph{emotion indicator}}, which detects the existence of emoticons and/or strong sentiment words in the documents. 
The strong sentiment words are hand-selected and represent words that are associated with a person's positive or negative feeling. 
We present some examples of strong sentiment words in Table~\ref{tbl:emoticons}, and provide the 
full list in the supplementary 
material made available online on the author's website.
We define $e$ to be $-1$ when only a negative emotion is observed and $e$ to be $1$ when only a positive emotion is observed, otherwise we treat $e$ as unobserved. 
Note that $e=0$ would correspond to a neutral emotion,
but we have no such observations so this is not considered.

The generative process of TOTM is as follows. 
First, we sample the document-aspect distribution $\theta_d$ for each document $d$,
\begin{align*}
\theta_d \sim \mbox{GEM}(\alpha^{\theta}, \beta^{\theta})~.
\end{align*}

Second, for $e=\{-1,1\}$, we model the emotion-sentiment distribution $\gamma_e$ by a Dirichlet distribution with asymmetric prior:
\begin{align*}
\gamma_e|e \sim \mbox{Dir}(\vec{q}_e)~.
\end{align*}
The prior $q_e$ is chosen such that $\vec{q}_{-1} = (0.9, 0.05, 0.05)$ and $\vec{q}_1 = (0.05, 0.05, 0.9)$.

Next, for the target words, we generate the aspect-target distribution $\psi_a$ for each aspect $a$:
\begin{align*}
\psi_a \sim \mbox{PYP}(\alpha^\psi, \beta^\psi, H_\psi)~.
\end{align*}
Here, $H_\psi$ is a discrete uniform vector over the vocabulary of the target words ($V_t$).

For the opinion words, we propose a novel hierarchical modeling that allows an opinion word to 
describe two different targets differently ({\it e.g.\ short} for {\it processing time} is good but 
{\it short} for {\it battery life} is bad), while at the same time allows for sharing of the 
polarity of opinion words between targets. 
This is achieved by assigning common base distributions to the target-opinion distributions. 
So target-opinion distributions $\phi'_{tr}$ for different targets $t$ share a common 
mean $\phi_r$ which itself is unknown so we sample it from a uniform
base $\phi^*_r$.
More specifically, for each $r=\{-1,0,1\}$ and $t=\{1,\dots,|V_t|\}$, we generate $\phi'_{tr}$ as follows:
%{
\vspace{-1mm}
\begin{align*}
\phi^*_r & = \vec{1/|V_o|}~, \\
\phi_r|\phi^*_r & \sim \mbox{PYP}(\alpha^\phi, \beta^\phi, \phi^*_r)~, \\
\phi'_{tr}|\phi_r & \sim \mbox{PYP}(\alpha^{\phi'}, \beta^{\phi'}, \phi_r)~,
\end{align*}
%}
where $V_o$ is the vocabulary of the opinion words.

Finally, for each target-opinion pair $\langle t_{dn},o_{dn}\rangle$ (indexed by $n$) in document $d$, we sample the respective aspect $a_{dn}$, sentiment $r_{dn}$ and the target-opinion pair:
%{
\vspace{-1mm}
\begin{align*}
a_{dn}|\theta_d & \sim \mbox{Discrete}(\theta_d)~, \\
r_{dn}|e_d,\gamma & \sim \mbox{Discrete}(\gamma_{e_d})~, \\
t_{dn}|a_{dn},\psi & \sim \mbox{Discrete}(\psi_{a_{dn}})~, \\
o_{dn}|t_{dn},r_{dn},\phi' & \sim \mbox{Discrete}(\phi'_{t_{dn},r_{dn}})~.
\end{align*}
%}

We note that each PYP distribution is parameterized by its own set of hyperparameters, {\it i.e.}\ $\beta^\theta$ differs for different document $d$, albeit not explicitly shown above for readability. 
We present a list of variables associated with TOTM in Table~\ref{tbl:variables}.
Also note that by modeling the target-opinion distribution explicitly, we have to store the information of the distribution for each target in the data, which is very large. 
In our implementation, we adopt a sparse representation for storing the counts associated with the target-opinion distributions. We find that each target word is only described by a limited number of opinion words in the data, which is less than 1\% of the words from the opinion word vocabulary.

In the next section, we propose a novel method to incorporate sentiment prior information for opinion analysis.

\section{Incorporating Sentiment Prior}
\label{sec:incorporatingPrior}

He \cite{He:2012:ISP:2184436.2184437} proposed a simple yet effective way to incorporate sentiment 
prior information into LDA by directly modifying the Dirichlet prior based on available sentiment lexicons. 
Naming her model LDA-DP (LDA with Dirichlet Prior modified), He replaces the topics in LDA by latent sentiment labels and allows the word priors to be custom probability distributions. 
The generative process of LDA-DP is identical to LDA and hence omitted in this paper.

In LDA-DP, the word distribution $\phi_r$ is Dirichlet distributed with the parameter 
$(\vec{\lambda}_r \times \alpha_r)$, where $r = \{-1, 0, 1\}$ is the sentiment label corresponding 
to negative, neutral and positive sentiment, respectively\footnote{We redefined the original 
sentiment labels~\cite{He:2012:ISP:2184436.2184437} for consistency.}. 
The $\lambda_{rv}$ is initialized to be $1/3$, and subsequently updated if the sentiment lexicon contains word $v$. 
In this case, $\lambda_{rv}$ takes the value of $0.9$ if the sentiment of word $v$ matches $r$, and takes the value of $0.05$ otherwise:
\begin{align*}
\lambda_{rv} =
\begin{cases}
0.9   & \mathrm{if\,\,\,} \mathrm{Sentiment}(v) = r \\
0.05  & \mathrm{otherwise}
\end{cases}
\end{align*}

Motivated by this, but not wishing to be required to give the exact strength by which the dictionary affects 
probabilities, 
instead, we propose a novel formulation that automatically learns and updates itself. 
We assume that a sentiment lexicon is available and provides sentiment scores for opinion words. 
Additionally, we assume that the sentiment score $S_v$ returned from the sentiment lexicon takes 
negative value when $v$ has negative sentiment, positive value when $v$ has positive sentiment, 
and $0$ when $v$ is neutral\footnote{We can simply normalize the score to conform to this assumption.}. 

Sentiment lexicons that are freely available online include 
SentiWordNet~\cite{baccianella2010sentiwordnet}, SentiStrength \cite{ASI:ASI21416}, 
MPQA Subjectivity lexicon~\cite{Wilson:2005:RCP:1220575.1220619} and others.
SentiStrength is developed from 
MySpace\footnote{MySpace is a social networking website similar to Facebook.} 
text data by a research group (Statistical Cybermetrics Research Group) from the 
University of Wolverhampton, UK. 
Since the SentiStrength lexicon is constructed for informal text, we use it to extract sentiment 
information for TOTM. 
The sentiment score $S_v$ from SentiStrenth ranges from $-5$ to $+5$, which conforms to our 
assumption. We assume that $S_v = 0$ for unlisted words. 

Additionally, we make use of the SentiWordNet 3.0 lexicon to evaluate TOTM. 
SentiWordNet is built on WordNet~\cite{fellbaum1999wordnet} by researchers from Italy.
We note that SentiStrength and SentiWordNet are developed independently by different teams
using different methods.
Thus we claim it is fair and unbiased to use
one lexicon for training and the other for evaluation.

Our formulation is as follows, introducing a tunable parameter $b$ that controls the strength of the prior, we replace the prior $\phi^*_r$ (in the context of TOTM) by the following:
\begin{align}
\phi^*_{rv} \propto (1+b)^{X_{rv}}~~, \label{eq:prior}
\end{align}
where $b>0$ and hence $\phi^*_{rv} > 0$. Here, $X_{rv}$ is the score of word $v$ for sentiment $r$, which is defined as
\begin{align*}
X_{rv} =
\begin{cases}
S_v     & \mathrm{if\ \ } r = 1 \mathrm{~~(positive)} \\
- |S_v| & \mathrm{if\ \ } r = 0 \mathrm{~~(neutral)} \\
- S_v   & \mathrm{if\ \ } r = -1 \mathrm{~~(negative)}~~~.
\end{cases}
\end{align*}
%We name our prior formulation of Equation~\ref{eq:prior} the power sentiment prior (PSP) formulation. 
Note that although there are multiple ways to formulate the prior, we choose the above formulation 
due to its simplicity and intuitiveness. 
We can see that positive $X_{rv}$ boosts the probability of word $v$ while a negative $X_{rv}$ diminishes it. 
Also, this formulation ensures the positivity of the prior, which can be difficult to achieve if we 
use other formulations such as a polynomial function.

Even though $b$ is a tunable parameter, we do not need to manually tune it. 
We propose a flexible way to learn the parameter $b$ from its posterior distribution (detailed in Subsection~\ref{subsec:hyperparameter}), thus relieving us from choosing the value for $b$, which can be difficult (the value of $b$ should depend on the sentiment score of the lexicon).

%We also apply our PSP formulation to LDA-DP and ILDA by modifying the prior of $\phi_r$, {\it i.e.}\ by replacing $\vec{\lambda}_r \times \alpha_r$ in LDA-DP and $\alpha^\phi$ in ILDA with $\phi^*_r$. As we will see later in the experiment section, using our PSP formulation leads to significant improvement in the sentiment evaluation.

\vspace{-2mm}
\section{Inference Technique}
\label{sec:modellikelihood}

In this section, we discuss the collapsed Gibbs sampler for TOTM, and then discuss the
sampling of the hyperparameters.
%We expand on the representation in the supplementary material.

\subsection{Collapsed Gibbs Sampling for TOTM}

\begin{algorithm}[t!]
\caption{Collapsed Gibbs Sampling for TOTM}
\label{alg:gibbs}
\begin{enumerate}[itemindent=-10pt, itemsep=1pt]
\item \parbox[t]{\dimexpr\linewidth+10pt}{
Initialize the model by assigning a random aspect to each target-opinion pair, sampling the sentiment label, and building the relevant customer counts $c^\mathcal{N}_k$ and table counts ${c'}^\mathcal{N}_k$ for all nodes.
}
\item For each document $d$:
\begin{enumerate}[itemindent=-10pt, noitemsep, nolistsep]
  \item For each target phrase $t_{dn}$:
  \begin{enumerate}[itemindent=-10pt, noitemsep, nolistsep]
    \item Decrement counts associated with $t_{dn}$.
    \item Sample new aspect $a_{dn}$ and corresponding parts of 
$\mathbf{C}$ from Equation~\ref{eq:conditional_posterior}.
    \item Increment associated counts for the new $a_{dn}$.
  \end{enumerate}
  \item For each opinion phrase $o_{dn}$:
  \begin{enumerate}[itemindent=-10pt, noitemsep, nolistsep]
    \item Decrement counts associated with $o_{dn}$.
    \item Sample new sentiment $r_{dn}$ and corresponding parts of 
$\mathbf{C}$ 
(like Equation~\ref{eq:conditional_posterior}).
    \item Increment associated counts for the new $r_{dn}$.
  \end{enumerate}
\end{enumerate}
\item \parbox[t]{\dimexpr\linewidth+10pt}{
Repeat step 2 until the model converges or when a fixed number of iterations is reached.
}
\end{enumerate}
\vspace{-1mm}
\end{algorithm}

The key to Gibbs sampling with PYPs is to marginalize out the probability vectors 
({\it e.g.}\ $\theta$) in the model and record various associated counts instead,
thus yielding a collapsed sampler.
While a common approach here is to use the Chinese Restaurant Process (CRP) representation 
of Teh and Jordan~\cite{teh2010hierarchical}, we use another representation that requires no 
dynamic memory and has better inference efficiency~\cite{chen2011sampling}. 
%This is more easily considered as a hierarchical extension to
%the standard Dirichlet distribution.
We let $g(\mathcal{N})$ be the marginalized 
likelihood associated with the probability vector $\mathcal{N}$.
The vector is marginalized out,
thus the likelihood is in terms of --- using the CRP terminology --- the {\it customer counts} $\mathbf{c}^\mathcal{N} = (\dots, c_i^\mathcal{N}, \dots)$
and the total customer count $C^\mathcal{N}$ (the sum of $c^\mathcal{N}_i$).
For the PYP, we introduce the {\it table counts}
$\mathbf{c'}^\mathcal{N} = (\dots, {c'_i}^\mathcal{N}, \dots)$ that represents the subset of $\mathbf{c}^\mathcal{N}$
that gets passed up the hierarchy (as customer for the parent probability vector
of $\mathcal{N}$), and 
${C'}^\mathcal{N}$, the total table count.
For instance, looking at the sub-hierarchy in Figure~\ref{fig:totm}
for $\phi'_{tr} \leftarrow \phi_{r}  \leftarrow  \phi^*_{r}$,
the customer count $c^{\phi'_{tr}}_v$ for opinion index $v$ is associated with
the table count ${c'}^{\phi'_{tr}}_v$ which are
added to the customer count $c^{\phi_{r}}_v$ ($\phi_r$ is the parent of $\phi'_{tr}$).
The table count of $\phi_r$, ${c'}^{\phi_{r}}_v$, is in turn added to the customer count $c^{\phi^*_{r}}_v$.
%Each time a smaller number gets passed up the hierarchy as
Note that table count is always smaller than customer count (${c'}^\mathcal{N}_i\leq {c}^\mathcal{N}_i$).
These counts are latent, not observed, hence they are sampled during inference.

By using the above representation, we do not need to record the occupancy counts of each table, hence we do not need a dynamic storage.
% We provide a more comprehensive explanation in the supplementary material.
The marginalized likelihood is given by
%\vspace{-0.5mm}
\begin{align}
g(\mathcal{N}) = \frac{(\beta^\mathcal{N}|\alpha^\mathcal{N})_{{C'}^\mathcal{N}}}{(\beta^\mathcal{N})_{C^\mathcal{N}}} \prod_i S^{c^\mathcal{N}_i}_{{c'}^\mathcal{N}_i, \alpha^\mathcal{N}}~~~,   \label{eq:modularized_likelihood}
\end{align}
\noindent
where $S^x_{y,\alpha}$ is the generalized Stirling number, whereas $(x)_C$ and $(x|y)_C$ denote the Pochhammer symbol~\cite{buntine2012bayesian}.
%For those familiar with CRP terminology, the associated count
%${c'}^\mathcal{N}_i$ corresponds to the number of tables open
%for data type $i$ and the advantage over CRP sampling is that one
%does not need to know the occupancy counts at tables
%(which requires dynamic storage since the number of tables is
% unknown).

We use bold face capital letters to denote the set of all relevant lower case variables, 
{\it e.g.}\ $\mathbf{A} = \{\vec{a}_1,\cdots,\vec{a}_D\}$,
where each $\vec{a}_i = \{a_{i1},\cdots,a_{i,N_d}\}$, denotes the set of all aspects.
Variables $\mathbf{R}, \mathbf{T}$ and $\mathbf{O}$ are defined similarly. 
In addition, we denote $\mathbf{C}$ to be the set of the customer counts and table counts for all probability vectors ($c^{\phi'_{tr}}, c^{\phi_{r}}, c^{\phi^*_{r}}$, {\it etc.})
Also, we denote $\mathbf{\zeta}$ the set of all hyperparameters (such as the $\alpha$'s). 
Note all probability vectors are marginalized out.
The likelihood of the model can then be written --- in terms of $g(\cdot)$ --- as 
$p(\mathbf{A}, \mathbf{R}, \mathbf{T}, \mathbf{O}, \mathbf{C} | \mathbf{\zeta}) \propto$
{
\setlength{\belowdisplayskip}{0pt}
\begin{align}
% p(\mathbf{A}, \mathbf{R}, \mathbf{T}, \mathbf{O}, \mathbf{C} | \mathbf{\zeta}) \propto
\Big(\!\prod_{d=1}^D g(\theta_d)\Big) \Big(\!\!\!\!\!\!\!\!\!\prod_{\ \ e=\{-1,1\}} \!\!\!\!\!\!\!\! g(\gamma_e)\Big) 
\Big(\!\prod_{a=1}^A g(\psi_a)\Big)
\Big(\!\!\prod_{r=-1}^1 \! g(\phi_r) 
\Big(\!\prod_{t=1}^{|V_t|} g(\phi'_{tr})\Big)
%\Big(\!\prod_{o=1}^{|V_o|} \big(\frac{1}{|V_o|}\big)^{{c'_o}^{\phi_r}} \Big)
\Big) 
%\Big(\!\prod_{t=1}^{|V_t|} \big(\frac{1}{|V_t|}\big)^{\sum_a {c'_t}^{\psi_{a}}} \Big)
. \label{eq:likelihood}
\end{align}
}

We use the collapsed Gibbs sampler from Chen et al.~\cite{chen2011sampling} for inference.
The concept of the sampler is analogous to LDA, which consists of decrementing counts 
associated with a word, sampling the respective new latent values for the word, and 
incrementing the respective counts. 
In our case, the process is more complicated, albeit following the same general procedure.
For the decrementing procedure, the table counts are represented as a
sum of Bernoulli ``indicator'' variables $u$. Each data item (customer) corresponding to 
a $+1$ in ${c}^\mathcal{N}_i$ either has $u=0$ or $u=1$.
When $u=1$, the data item is passed up the hierarchy to the
parent of $\mathcal{N}$, and thus
contributes a $+1$ to the table count ${c'}^\mathcal{N}_i$.
% As we do not store the indicator $u$ explicitly, we sample them and decrement the 
%respective counts accordingly \highlight{(detail needed?)}. 
% See supplementary material for a detailed procedure.
Note that the counts can only increase or decrease by one, since we are decrementing
and incrementing a word at a time.

%Chen et al.'s sampler then does not sample ${c'}^\mathcal{N}_i$ 
%but rather each $u$, and thus at any point during the
%sampling the counts can only increment by at most $1$.

When sampling a new aspect $a$ or sentiment $r$, the modularized likelihood  
(Equation~\ref{eq:likelihood})
allows the posterior to be
computed quickly, since the conditional posterior simplifies to a ratio of likelihoods.
This in turn allows for the ratio to simplify further since the counts can only
change by 1. 
For instance, the ratio of the Pochhammer symbols, $(x|y)_{C+1} / (x|y)_C$, is reduced to a constant. 
While for the ratio of Stirling numbers, such as 
$S^{y+1}_{x+1, \alpha}/S^{y}_{x, \alpha}$,
can be computed quickly via caching~\cite{buntine2012bayesian}.
%The full conditional posterior probability for Gibbs sampling is just a ratio of the posterior distributions.

For example, the conditional posterior for aspect $a_{dn}$ is
\begin{align}
p(a_{dn}, \mathbf{C} | \mathbf{A}^{-dn}, \mathbf{R}, \mathbf{T}, \mathbf{O}, \mathbf{C}^{-dn}, \mathbf{\zeta}) & \nonumber \\
= \frac{p(\mathbf{A}, \mathbf{R}, \mathbf{T}, \mathbf{O}, \mathbf{C} | \mathbf{\zeta})}{p(\mathbf{A}^{-dn}, \mathbf{R}, \mathbf{T}, \mathbf{O}, \mathbf{C}^{-dn} | \mathbf{\zeta})} & ~~~, \label{eq:conditional_posterior}
\end{align}
where the superscript $\Box^{-dn}$ indicates that the target-opinion pair $\langle t_{dn}, a_{dn} \rangle$ is removed from the respective sets.
It is trivial to show that the conditional posterior simplifies to ratios of Pochhammer symbols and 
a ratio of Stirling numbers with Equation~\ref{eq:modularized_likelihood} and 
Equation~\ref{eq:likelihood}.
The conditional posterior probability for sampling the sentiment $r_{dn}$ can be similarly written.
%\begin{align}
%p(r_{dn}, \mathbf{C} | \mathbf{A}, \mathbf{R}^{-dn}, \mathbf{T}, \mathbf{O}, \mathbf{C}^{-dn}, \mathbf{\zeta}) & \nonumber \\
%= \frac{p(\mathbf{A}, \mathbf{R}, \mathbf{T}, \mathbf{O}, \mathbf{C} | \mathbf{\zeta})}{p(\mathbf{A}^, \mathbf{R}^{-dn}, \mathbf{T}, \mathbf{O}, \mathbf{C}^{-dn} | \mathbf{\zeta})} & ~~~.\label{eq:conditional_posterior2}
%\end{align}

Note the change in associated counts $\mathbf{C}|\mathbf{C}^{-dn}$
will be the full possible range of $+1$'s propagated up the
hierarchy.  So sampling $r_{dn}=r$ will
increment $c^{\phi'_{t_{dn}r}}_{o_{dn}}$ and may/may-not
increment ${c'}^{\phi'_{t_{dn}r}}_{o_{dn}}\!.$
If it does increment ${c'}^{\phi'_{t_{dn}r}}_{o_{dn}}$ then it also increments
$c^{\phi_{r}}_{o_{dn}}$,
but then ${c'}^{\phi_{r}}_{o_{dn}}$ may or may-not be incremented.
Sampling all these increments corresponds to sampling on a small
tree of Booleans which can be done in closed form.
Similarly, sampling a new $a_{dn}=a$ will increment $c^{\theta_d}_a$, and if 
${c'}^{\theta_d}_a$ is also incremented, a new aspect cluster is created for $t_{dn}$.

We summarize the collapsed Gibbs sampler in Algorithm~\ref{alg:gibbs}, and refer the interested reader to the supplementary material for detail.

\subsection{Hyperparameters Sampling}
\label{subsec:hyperparameter}

During inference, we sample the hyperparameters of the PYP using an auxiliary variable sampler \cite{teh2006bayesian}. 
Moreover, we propose a novel method to update the hyperparameter $b$, which controls the strength of the sentiment prior. 
Instead of sampling the hyperparameter $b$ ({\it e.g.}\ using the slice sampler~\cite{Neal:AS03}), we adopt an optimization approach since the posterior of $b$ is highly concentrated in a small region (thin-tailed). 
The posterior density is given by the following equation, subject to a normalization constant.
\begin{align*}
p(b|\vec{c}) \propto p(b) \prod_{r} \prod_v \left( \frac{(1+b)^{X_{rv}}}{\sum_{i} \sum_j (1+b)^{X_{ij}}} \right)^{c_{rv}}~,
\end{align*}
where ${c_{rv}}$ is the number of times a word $v$ is assigned a sentiment $r$, and $p(b)$ is the hyperprior of $b$.
%\footnote{Technically, $c_{rv} = {c'}^{\phi_r}_v$.}
We assume a weak hyperprior for $b$, 
$b \sim \mathrm{Gamma}(1,1)$.

During inference, we update $b$ to its {\it maximum a posteriori probability} (MAP) estimate using a gradient ascent algorithm. 
We optimize the log posterior $l(b) = \log(p(b|\vec{c}))$ since $\log$ is an increasing function. 
The gradient of the log posterior is derived as
\begin{align*}
l'(b) = \frac{1}{(1+b)} \sum_r \sum_v c_{rv} \left( X_{rv} - \mathbb{E}_{\phi_r}[X_r] \right) + \rho'(b)~~,
\end{align*}
where $\mathbb{E}_{\phi_r}[X_r]$ is the expected value of $X_r$ under the probability distribution $\phi_r$, and $\rho'(b)$ is the derivative of $\log p(b)$. 
We summarize the gradient ascent algorithm in Algorithm~\ref{alg:gradient_ascent}.
Additionally, in the supplementary material, we present the gradient derivation and a plot of the log posteriors of $b$ given different statistics $\vec{c}$.

\begin{algorithm}[t]
\caption{Gradient Ascent Optimization for Hyperparameter $b$}
\label{alg:gradient_ascent}
\begin{enumerate}[itemindent=-10pt, itemsep=1pt]
\item \parbox[t]{\dimexpr\linewidth+10pt}{
Given an initial value for $b = b_0$, evaluate the gradient $l'(b_0)$.
}
\item \parbox[t]{\dimexpr\linewidth+10pt}{Given a learning rate $\tau$, update $b$ to $b_i = b_{i-1} + \tau \times l'(b_{i-1})$, if the new log posterior $l(b_i)$ is lower than $l(b_{i-1})$, we halve the learning rate: $\tau := \tau/2$~.}
\item \parbox[t]{\dimexpr\linewidth+10pt}{Repeat step 2 until $b$ converges.}
\end{enumerate}
\vspace{-2mm}
\end{algorithm}
\vspace{-1mm}

\section{Data}
\label{sec:data}

%\subsubsection{Tweets}

For experiments, we perform aspect-based opinion analysis on tweets, which are characterized by their limited 140 characters text. 
From the {\it Twitter 7} 
dataset\footnote{\small \url{http://snap.stanford.edu/data/twitter7.html}} 
\cite{yang2011patterns}, we queried for tweets that are related to electronic products such as {\it camera} and {\it mobile phones} (see the list of our query words in the supplementary material). 
We then remove non-English tweets with {\it langid.py}~\cite{lui2012langid}.
Moreover, since most spam tweets contain a URL, we adopt a conservative approach to remove spam by 
discarding tweets containing URLs. 
This results in a dataset of about 9 million tweets, which
we name as the electronic product dataset.

Due to the lack of sentiment labels on the electronic product dataset, we make use of the 
Sentiment140 (Sent140) tweets\footnote{\small \url{http://help.sentiment140.com/home}}~\cite{go2009twitter} 
for sentiment classification evaluation.
Each Sent140 tweet contains a sentiment label (positive or negative) that are determined by emoticons. 
The whole corpus contains 1.6 million tweets, with half of them labeled as positive and the other half as negative.

In addition, we also use the SemEval 2013 dataset\footnote{
%\resizebox{1.1\linewidth}{!}{{\parbox{\linewidth}
{\small \url{http://www.cs.york.ac.uk/semeval-2013/task2/}}
%}}
}~\cite{Nakov_semeval-2013task} for evaluation.
SemEval tweets are annotated on Mechanical Turk, which arguably provides better sentiment labels compared to Sent140.
Since annotation is expensive, SemEval has only 6322 tweets.
%Although emoticons were explicitly removed in Sent140 tweets, we treat the sentiment labels as seen emoticons during training.

%\subsubsection{Review Data}
%
%In addition to tweets, we perform experiments with the popular Movie Review Data\footnote{\url{http://www.cs.cornell.edu/people/pabo/movie-review-data/} (polarity v2.0)} \cite{Pang:2004:SES:1218955.1218990}. This dataset contains 1000 positive and 1000 negative movie reviews obtained from IMDB.

\subsection{Data Preprocessing}
\label{subsec:preprocessing}

\hspace{-0.3mm}Here, we describe the preprocessing steps that we apply to tweets. 
Firstly, we apply Twitter NLP~\cite{owoputi2013improved}, a state-of-the-art tool for part-of-speech (POS) tagging on tweets.
We then apply word normalization to clean up the tweets.
We make use of the lexical normalization 
dictionary\footnote{\small \url{http://ww2.cs.mu.oz.au/~tim/\#resources}} 
from Han et al.~\cite{Han:2012:ACN:2390948.2391000}, 
but modify it such that proper nouns are not normalized. 
For instance, words like `iphone' and `xbox' are not normalized, since they are the targets we are interested in.
We perform normalization after POS tagging since tweets normalization degrades the performance of Twitter NLP~\cite{Han:2013:LNS:2414425.2414430}.

Next, we proceed to extract target-opinion pairs from the data.
Following Moghaddam and Ester~\cite{Moghaddam:2012:DLM:2396761.2396863}, we apply the Stanford Dependency Parser~\cite{de2006generating} to extract dependency relations that will be used to form the target-opinion pairs. 
However, our approach is slightly different: we do not use the {\it Direct Object} ({\it dobj}) relation to obtain a target-opinion pair, for example, the sentence ``{\it I like the perfect picture quality}'' gives `{\it dobj}(like, picture quality)' and `{\it amod}(picture quality, perfect)', resulting in two target-opinion pairs, $\langle${\it picture quality, like}$\rangle$ and $\langle${\it picture quality, perfect}$\rangle$. 
We drop the target-opinion pair associated with {\it dobj} and instead use the {\it dobj} relation for the emotion indicator variable. 
Note that we use the {\it caseless English model} in the Stanford Dependency Parser, which works better for tweets. 
Additionally, since standard NLP tools perform less optimally on 
tweets~\cite{Ritter:2011:NER:2145432.2145595}, we use the POS tagging from Twitter NLP to clean up 
the target-opinion pairs.
We note that negations like `{\it not}' are captured as dependency relations, the negated words are then treated as new words with the prefix `{\it not\_}'.

We determine the emotion indicator variable {\it via} the existence of emoticons, strong sentiment words and/or the {\it dobj} relation in each tweet. 
We simply set the emotion indicator to $-1$ (negative) or $1$ (positive) as long as the 
indicators agree with one another, and unobserved otherwise.
The list of emoticons used is compiled from 
Wikipedia\footnote{{\small \url{http://en.wikipedia.org/wiki/Kaomoji}} and \\ 
%\resizebox{1.09\linewidth}{!}{{\parbox{\linewidth}
{\small \url{http://en.wikipedia.org/wiki/List_of_emoticons}}
%}}
}.
We present a subset of the emoticons and strong sentiment words in Table~\ref{tbl:emoticons}, while the full list is available in the supplementary material.
For Sent140 and SemEval tweets, we replace the unobserved emotion indicator by their sentiment label.

\begin{table}[tb!]
	\centering
	\vskip -2mm
    \caption{Emoticons and Strong Sentiment Words}
    \label{tbl:emoticons}
    \vskip 0.15in
	\begin{tabular}{|p{0.41\columnwidth}|p{0.41\columnwidth}|}
    \hline
	\multicolumn{1}{|c|}{Positive} & \multicolumn{1}{c|}{Negative} \\
	\hline
	{\tt 
	:-) :o) :] :3 :c) :> =] 8) =) :\}
	:-D ;-D :D 8-D
	\textbackslash o/ 
	$^\wedge$\_$^\wedge$ ($^\wedge$O$^\wedge$)/ ($^\wedge\!\,^\wedge$)/
	:\}
	} & {\tt
	>:-( >:[ :-( :c
	:@ >:( ;( ;-(
	:'-( :'(
	D;
	(T\_T) (;\_;) (;\_:)
	T.T !\_!
	} \\
	\hline
	{\tt happy glad love delighted like } & {\tt sad upset hate dislike angry} \\
	\hline
	\end{tabular}
	\vskip -0.1in
\end{table}

We then perform tweet aggregation, which is found to give significant improvement for LDA~\cite{Mehrotra:2013:ILT:2484028.2484166}. 
We group tweets that contain the same hashtag (word prefixed with \# symbol) or same mention (word prefixed with @ symbol) into a single document, this allows co-occurrence within the same {\it tags} (our abbreviation for hashtags and mention) to be used by topic models. 
Grouping tweets also allows us to summarize the results for each tag, giving us a better opinion overview (see Subsection~\ref{subsec:qualitative} for example).
Additionally, we discard tags that occur infrequently.
We note that although tweets are merged to form a larger document, the emotion indicator (variable $e$) is observed and stored for each individual tweet (rather than the merged document), this prevents the emotion indicator from being lost through merging.

\begin{figure}[b!]
\vskip -0.1in
\begin{center}
\centerline{\includegraphics[width=0.735\columnwidth]{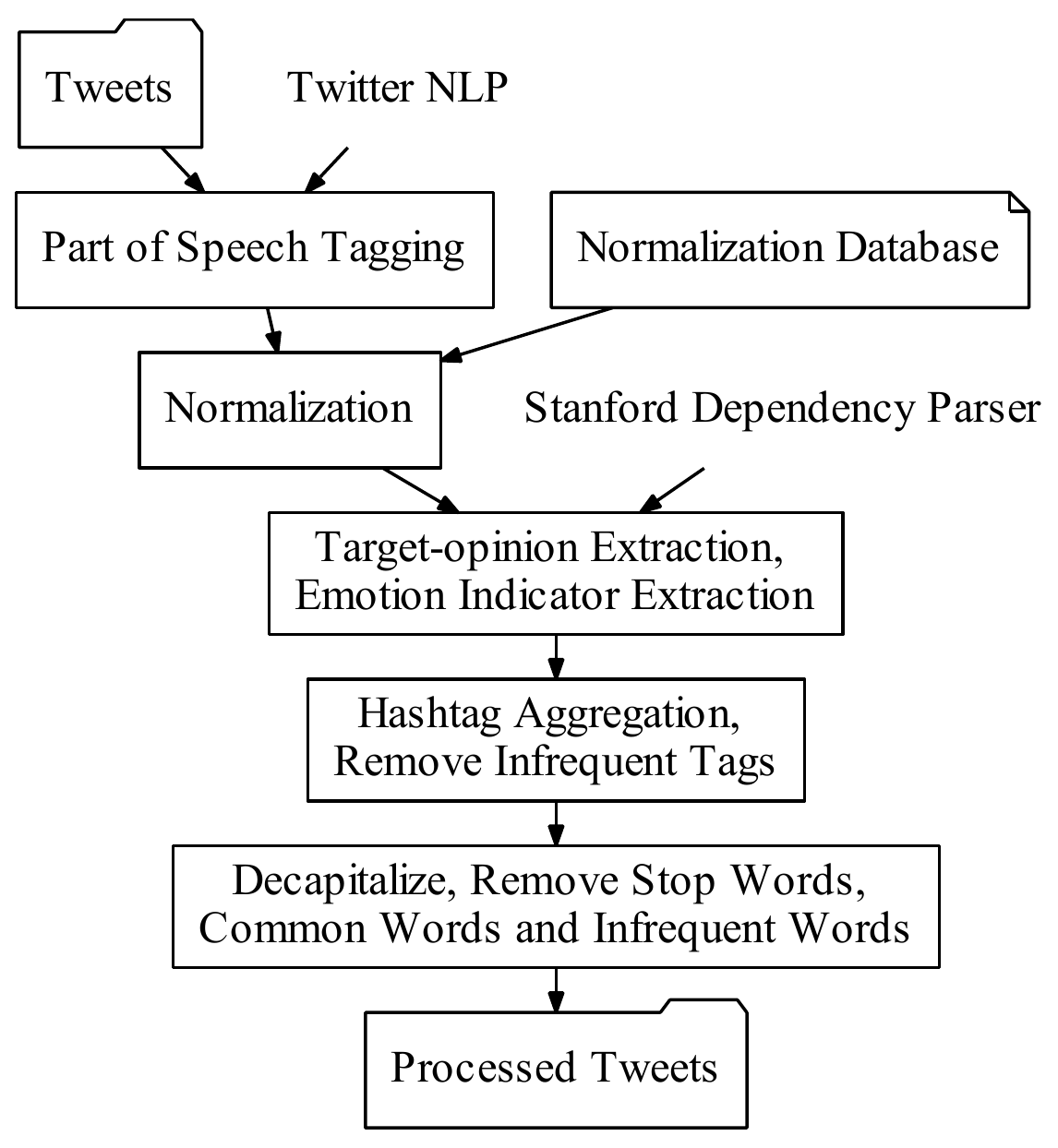}}
\caption{Preprocessing Pipeline}
\label{fig:preprocessing}
\end{center}
\vskip -0.25in
%\vspace{-3mm}
\end{figure}

Finally, we perform other standard preprocessing techniques to topic modeling, this consists 
of decapitalizing the words, removing stop words and discarding commonly occurred words and infrequent 
words.
%words\footnote{Infrequent words are not removed for SemEval data due to size.}.
We define the common words as words that appear in at least 90\% of the documents, and infrequent words as words that appear less than 50 times in the corpus.
We randomly split the data into 90\% training set and 10\% test set for evaluation. 
We present a summary of the preprocessing pipeline in Figure~\ref{fig:preprocessing}. 

\subsection{Corpus Statistics}

On average, we found that there are $0.69$ target-opinion pair extracted per electronic product tweet. 
Out of the electronic tweets that contain at least one target-opinion pair, 17.9\% of them contain an emotion indicator. After preprocessing, the number of unique target word tokens in the electronic product tweets is 4402, while the number of unique opinion word tokens is 25188. 
We present a summary of the corpus statistics for all datasets in Table~\ref{tbl:corpusStats}.

For the electronic product tweets, the top tags are \#apple, \#phone, \#iphone, \#computer and \#laptop. 
We note that some tags are associated with products, brands or companies, for example, \#playstation and \#xbox are associated with gaming products, while \#sony and \#canon are associated with companies. 
In Subsection~\ref{subsec:qualitative} below, we show that aggregating hashtags allow us to have a more focused view on certain products or companies, as well as facilitating comparison between these products or companies side-by-side.

\begin{table}[tb!]
	\centering
	\vskip -2mm
    \caption{Corpus Statistics}
    \label{tbl:corpusStats}
    \vskip 0.15in
	\begin{tabular}{|l|c|c|c|}
    \hline
	& Electronic & Sent140 & SemEval \\
	\hline
	Number of tweets & $\sim$9M & 1.6M & 6322 \\
	\hline
	Opinion pairs per tweet & 0.69 & 0.41 & 0.47 \\
	\hline
	\% tweets containing $e$ & 17.9 & 100 & 57.5 \\
	\hline
	Target vocabulary & 4402 & 1050 & 1875 \\
	\hline
	Opinion vocabulary & 25188 & 8599 & 813 \\
	\hline
	\end{tabular}
	\vskip -0.11in
\end{table}

\section{Experiments and Results}
\label{sec:experiment}

In this section, we demonstrate the usefulness of TOTM for opinion mining. 
We evaluate TOTM quantitatively against ILDA and LDA-DP in terms of perplexity and sentiment classification.
To compare the effectiveness of various sentiment lexicons, we propose a novel sentiment metric to evaluate the sentiment-opinion word distributions $\phi$'s.
Qualitatively, we utilize TOTM for the task of opinion mining from the electronic product tweets, and show that we are able to extract various useful opinions on technological products such as iPhone.

\subsection{Experiment Settings}

For all the experiments, we initialize the hyperparameters of PYP to $\alpha=\beta=0.1$ and the sentiment hyperparameter to $b=10$, noting that the hyperparameters are optimized automatically as discussed in Subsection~\ref{subsec:hyperparameter}.

To determine the optimal number of latent aspects ($A$) for ILDA, we set aside 5\% of the training data as development set, and select $A$ (tested in increment of 10) such that perplexity of the development set is minimized. 
For a fair comparison between TOTM and ILDA, we cap the maximum number of aspects of TOTM to be that of ILDA. 
Our experiment finds that the number of aspects in TOTM always converges to the cap.
%, although it can be smaller. 
We note that LDA-DP has only three fixed `{\it topics}', which is the number of sentiments.

During inference, we run the collapsed Gibbs algorithm until the convergence criteria is satisfied, 
defined by which the training log likelihood does not differ by more than 0.1\% in ten 
consecutive iterations. 
Empirically, we find that all experiments converge within 200 iterations, indicating a good Gibbs sampling algorithm.

\subsection{Quantitative Evaluations}

\subsubsection{Perplexity}

We compute the perplexity of the test set to measure how well the models fit to the data. 
The perplexity is negatively related to the likelihood of the test data. 
Since aspect-based opinion analysis deals with two types of vocabulary, we compute the perplexity for both target words and opinion words, in this case:
{
\setlength{\abovedisplayskip}{3pt}
\begin{align*}
\mathrm{perplexity}(\mathbf{W}) = \exp\left(-\frac{\sum_{d=1}^D \log P(\vec{w}_d)}{\sum_{d=1}^D N_d}\right)~,
\end{align*}
}where $\mathbf{W}$ can be either target words $\mathbf{T}$ or opinion words $\mathbf{O}$, $N_d$ is the number of the target-opinion pairs in document $d$. We also compute the overall perplexity, which is given by
{
\setlength{\abovedisplayskip}{2pt}
\setlength{\belowdisplayskip}{2pt}
\begin{align*}
\mathrm{perplexity}(\mathbf{T}, \mathbf{O}) = \exp\left(-\frac{\sum_{d=1}^D \log P(\vec{t}_d, \vec{o}_d)}{2 \sum_{d=1}^D N_d}\right)~.
\end{align*}
}

We present the perplexity result (the lower the better) for the electronic product tweets in Table~\ref{tbl:perplexityTweet}. 
We present the perplexity result of Sent140 tweets and SemEval tweets in the supplementary material, for which the same conclusion can be drawn.
From the perplexity results, it is clear that modeling the target-opinion pairs directly leads to significant improvement of opinion words perplexity and hence the overall perplexity.
Note that LDA-DP only models the opinion words, thus we can only compare the perplexity for opinion words, we can see that its result is comparable to that of ILDA, albeit slightly better.

\begin{table}[tb!]
	\centering
	\vskip -2mm
    \caption{Test Perplexity on Electronic Product Tweets}
    \label{tbl:perplexityTweet}
	\vskip 0.15in
%    \resizebox{0.9\linewidth}{!}
    {
	\begin{tabular}{|l|r@{\ \tiny$\pm$\ }l|r@{\ \tiny$\pm$\ }l|r@{\ \tiny$\pm$\ }l|}
    \hline
	& \multicolumn{2}{c|}{Target} & \multicolumn{2}{c|}{Opinion} & \multicolumn{2}{c|}{Overall} \\
	\hline
	LDA-DP & \NAcell & $ 510.15 $ & {\tiny $0.08$} & \NAcell \\
	\hline
	ILDA & $ 594.81 $ & {\tiny $ 13.61 $} & $ 519.84 $ & {\tiny $ 0.43 $} & $ 556.03 $ & {\tiny $ 6.22 $} \\ 
	\hline
	TOTM & $ 592.91 $ & {\tiny $ 13.86 $} & $ \mathbf{ 137.42 } $ & {\tiny $ 0.28 $} & $ \mathbf{ 285.42 } $ & {\tiny $ 3.23 $} \\
	\hline
	\end{tabular}
	}
	\vskip -0.01in
\end{table}

\subsubsection{Sentiment Classification}

% % Single column table
\begin{table}[tb!]
	\centering
	\vskip -2mm
    \caption{Sentiment Classification Results (\%)}
    \label{tbl:classification}
	\vskip 0.15in
    \resizebox{1\linewidth}{!} 
    {
	\begin{tabular}{|l|c|c|c|c|}
    \hline
    {\it Sent140 Tweets} & Accuracy & Precision & Recall & F1-score \\
    \hline
	LDA-DP & $57.3$ & $56.1$ & $90.1$ & $69.2$ \\
	\hline
	ILDA & $54.1$ & $56.9$ & $55.3$ & $55.9$ \\
	\hline
	TOTM & $\mathbf{65.0}$ & $\mathbf{61.7}$ & $\mathbf{90.2}$ & $\mathbf{73.3}$ \\
	\hline \hline
	{\it SemEval Tweets} & Accuracy & Precision & Recall & F1-score \\
	\hline
	LDA-DP & $ 52.1 $ & $ 65.0 $ & $ 58.3 $ & $ 61.4 $ \\
	\hline
	ILDA & $ 46.8 $ & $ 60.7 $ & $ 53.6 $ & $ 56.3 $ \\
	\hline
	TOTM & $ \mathbf{73.3 }$ & $ \mathbf{84.0 }$ & $ \mathbf{74.9 }$ & $ \mathbf{79.0 }$ \\
	\hline
	\end{tabular}
	}
	\vskip -0.1in
\end{table}

\begin{table*}[t!]
    \centering
    \vskip -2mm
    \caption{Sentiment Evaluations for the Sentiment Priors (in unit of $\mathbf{0.01}$)}
    \label{tbl:sentimentEvaluationAll}
    \vskip 0.15in
    \resizebox{0.83\linewidth}{!}
    {
    \begin{tabular}{|l|r@{\ \tiny$\pm$\ }l|r@{\ \tiny$\pm$\ }l|r@{\ \tiny$\pm$\ }l|r@{\ \tiny$\pm$\ }l|r@{\ \tiny$\pm$\ }l|r@{\ \tiny$\pm$\ }l|}
    \hline
     & \multicolumn{4}{c|}{{\it Electronic Product Tweets}} & \multicolumn{4}{c|}{{\it Sent140 Tweets}} & \multicolumn{4}{c|}{{\it SemEval Tweets}} \\
    \hline
     & \multicolumn{2}{c|}{Negativity} & \multicolumn{2}{c|}{Positivity} & \multicolumn{2}{c|}{Negativity} & \multicolumn{2}{c|}{Positivity} & \multicolumn{2}{c|}{Negativity} & \multicolumn{2}{c|}{Positivity}  \\
    \hline
    No lexicon 
    & {$ 17.82 $} & {\tiny $1.26$} & {$ 17.39 $} & {\tiny $0.45$} 
    & {$ 22.63 $} & {\tiny $ 0.96 $} & {$ 32.31 $} & {\tiny $ 1.98 $} 
    & {$ 15.24 $} & {\tiny $ 1.45 $} & {$ 21.03 $} & {\tiny $ 3.85 $} \\
    \hline
    MPQA
    & {$ \mathbf{ 23.91 }$} & {\tiny $ 0.49 $} & {$ { 31.96 }$} & {\tiny $ 0.09 $} 
    & {$ 24.10 $} & {\tiny $ 0.49 $} & {$ \mathbf{ 42.65 }$} & {\tiny $ 1.02 $}
    & {$ 16.88 $} & {\tiny $ 0.31 $} & {$ 29.47 $} & {\tiny $ 0.99 $} \\
    \hline
    SentiStrength
    & {$ { 23.19 }$} & {\tiny $ 0.08 $} & {$ \mathbf{ 35.69 } $} & {\tiny $ 0.33 $} 
    & {$ \mathbf{ 24.29 }$} & {\tiny $ 1.07 $} & {$ { 41.26 } $} & {\tiny $ 1.53 $}
    & {$ \mathbf{ 16.94 }$} & {\tiny $ 0.78 $} & {$  \mathbf{ 32.17 }$} & {\tiny $ 2.07 $}  \\
    \hline
    \end{tabular}
    }
    \vspace{-0.1in}
\end{table*}

Here, we perform a classification task to predict the polarity of the test data for Sent140 and SemEval data. 
We determine the polarity of a test document $d$ by simply selecting the polarity $r$ that gives higher likelihood in $\phi_r$:
{
\setlength{\abovedisplayskip}{2pt}
\begin{align*}
\mathrm{polarity}(d) = \argmax_{r=\{-1,1\}} \prod_i \phi_{r, o_{di}}~~.
\end{align*}
}For simplicity, our evaluation is a binary classification task, as such, we do not include neutral tweets from SemEval data during evaluation. Note that Sent140 data does not have neutral tweets.

We present the classification {\it accuracy}, {\it precision}, {\it recall} and the {\it F1 score} in Table~\ref{tbl:classification}.
We can see that TOTM outperforms LDA-DP and ILDA on both datasets, suggesting that our prior formulation is more appropriate than that of LDA-DP.
We can also see that LDA-DP gives a better sentiment classification compared to ILDA, which does not incorporate any prior information.
Note that the classification result for SemEval data is better than that of Sent140.  We conjecture that this is because Sent140's sentiment labels are obtained from the emoticons, which are noisy in nature; while the sentiment labels for SemEval data is annotated.

\subsubsection{Evaluating the Sentiment Prior}

\begin{table}[t!]
	\centering
	\vskip -2mm
    \caption{Top Target Words for Electronic Product Tweets}
    \label{tbl:targetWords}
	\vskip 0.15in
    \resizebox{0.98\linewidth}{!} 
    {
	\begin{tabular}{|c|l|}
    \hline
    Aspects ($a$) & \multicolumn{1}{c|}{Target Words ($t$)} \\
	\hline
	Camera & camera, pictures, video camera, shots \\
	\hline
	Apple iPod & ipod, ipod touch, songs, song, music \\
	\hline
	Android phone & android, apps, app, phones, keyboard \\
	\hline
	Macbook & macbook, macbook pro, macbook air \\
	\hline
	Nintendo games & nintendo, games, game, gameboy \\
	\hline
	\end{tabular}
	}
	\vskip -0.135in
\end{table}

We propose a novel method to evaluate the learned sentiment-opinion phrase distributions $\phi$ by using another sentiment lexicon. 
We use the SentiWordNet lexicon for evaluation, noting that the lexicon used during training is the SentiStrength lexicon.

Unlike SentiStrength, the SentiWordNet lexicon provides two values for each word.
We name them the positive affinity $Z^+_v$ and negative affinity $Z^-_v$ for a given word $v$, they ranged from $0$ to $1$.
For example, the word `{\it active}' has a positive affinity of $0.5$ and a negative affinity of $0.125$; while `{\it supreme}' has a positive affinity or $0.75$ and a negative affinity of $0$.

Given the affinities, we propose the following sentiment score to evaluate an opinion word distribution $\phi_r$:
{
\setlength{\abovedisplayskip}{0pt}
\setlength{\belowdisplayskip}{1pt}
\begin{align*}
Score(\phi_r, Z) = E_{\phi_r}[Z] = \sum_{v=1}^{V_o} Z_v \phi_{rv}~~,
\end{align*}
}
where $Z$ is either $Z^+$ or $Z^-$, the positive or negative affinity. The sentiment score is also the expected sentiment under the opinion word distribution.

Here, we evaluate $\phi_{-1}$ with negative affinity $Z^-$ and $\phi_1$ with positive affinity $Z^+$. 
% Although we can also calculate the sentiment score with other combination ({\it e.g.}\ $\phi_1$ with $Z^-$), we find that they are less useful for comparison.
We compare the sentiment scores between the cases when a sentiment lexicon is used and when it is not.
Additionally, we also make use of the MPQA Subjectivity lexicon for sentiment prior (during training) and compare the sentiment evaluation against the SentiStrength lexicon.
We present the result in Table~\ref{tbl:sentimentEvaluationAll}.  As we can see, it is clear that incorporating prior information results in huge improvement in the sentiment score. 
Also, the priors for SentiStrength are slightly better than MPQA on average.
We note that optimizing the hyperparameter $b$ is very important, as it relieves us from tuning the hyperparameter manually.
To illustrate, the optimized $b$ converges to $2.59$ on the electronic product tweets, while on Sent140 and SemEval dataset, the $b$ converges to $1.85$ and $0.71$ respectively.
We also find that, in our tests, an incorrectly chosen $b$ can lead to a bad result.

%\newpage
\subsection{Qualitative Analysis and Applications}
\label{subsec:qualitative}

\subsubsection{Analyzing Word Distributions}

\begin{figure}[b!]
%\vskip 0.2in
\begin{center}
\vspace{-1mm}
%\centerline{\includegraphics[width=\columnwidth]{figures/heat_chart.jpg}}
\subfigure[Heat Map]{\includegraphics[height=4cm]{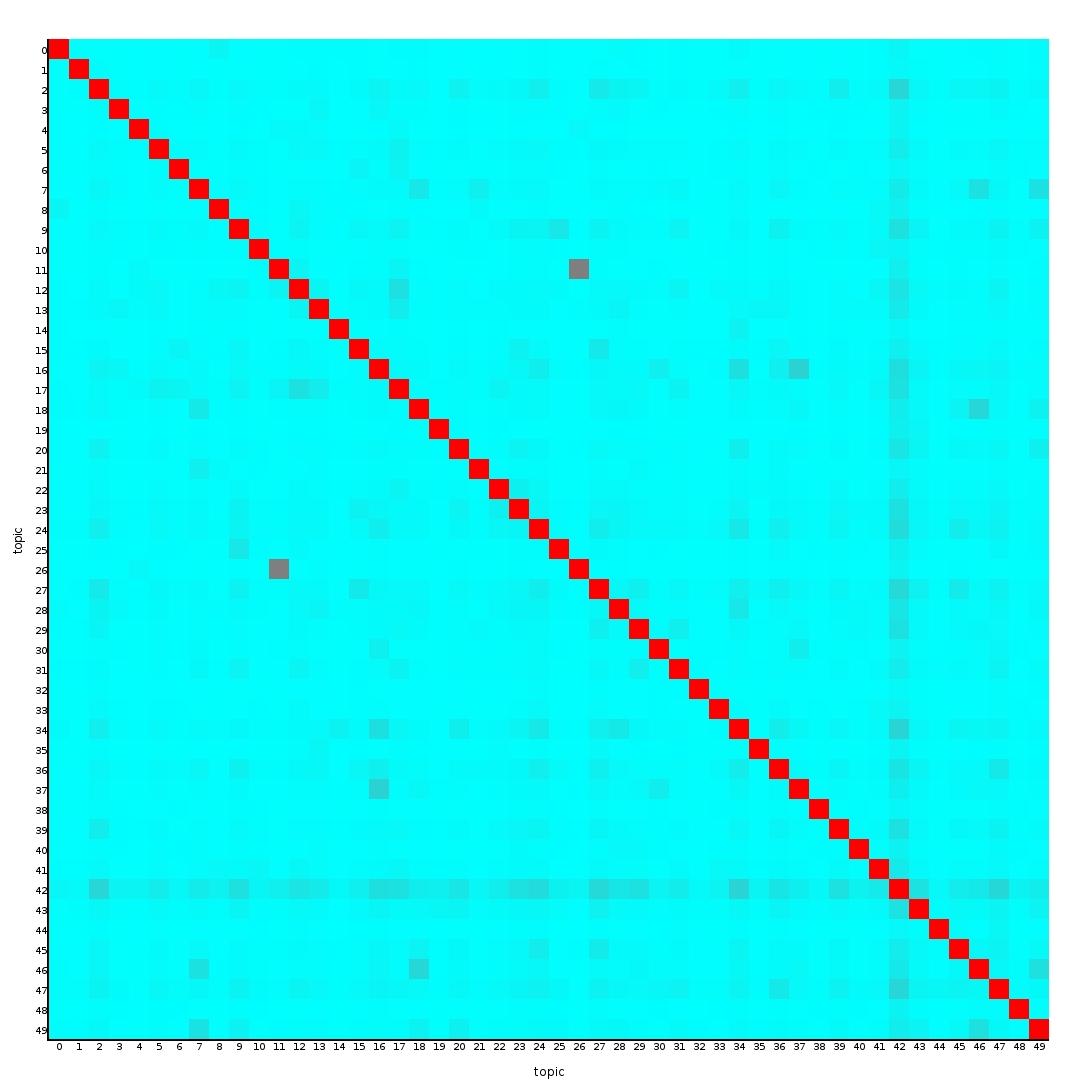}}
\subfigure[Legend]{\includegraphics[height=3.8cm]{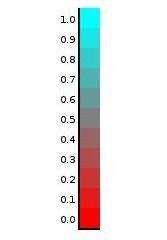}}
\caption{Pair-wise Hellinger Distances for Aspects (Colored)}
\label{fig:heat_map}
\vspace{-3.5mm}
\end{center}
%\vskip -0.2in
\end{figure}

First, we inspect the clustering of target words by TOTM and ILDA, noting that LDA-DP does not model the target words. 
We calculate the pair-wise Hellinger distance between each document-aspect distribution and found that the aspects are distinctive.
Hel\-lin\-ger distance is commonly used to measure the dissimilarity between two probability distributions.
%  for discrete distributions, the Hellinger distance between two probability distributions $\vec{p} = (p_1, \dots, p_K)$ and $\vec{q} = (q_1, \dots, q_K)$ is given as:
%  \begin{align*}
%  Hellinger(\vec{p},\vec{q}) = \frac{1}{\sqrt{2}} \left( \sum_{i=1}^K \left( \sqrt{p_i} - \sqrt{q_i} \right)^2 \right)^{\frac{1}{2}}~.
%  \end{align*}
The Hellinger distances between all pairs of aspect distributions from TOTM is displayed as a heat map in Figure~\ref{fig:heat_map}, we can see that the distances between the topics are high, indicating that there is no duplicated aspect. 
We note that the heat map for ILDA is similar and hence not presented here.
We also display an extract of the top target words from TOTM in Table~\ref{tbl:targetWords}.
Our empirical examination on the aspect-target word distributions suggest that both TOTM and ILDA perform well in clustering the target words.

\begin{table}[t!]
	\centering
	\vskip -2mm
    \caption{Opinion Analysis of Target Words with TOTM}
    \label{tbl:opinions}
	\vskip 0.15in
    \resizebox{0.98\linewidth}{!} 
    {
	\begin{tabular}{|c|c|l|}
    \hline
    Target ($t$) & $+/-$ & \multicolumn{1}{c|}{Opinions ($o$)} \\
	\hline
	\hline
	\multirow{2}{*}{phone} 
	& \textcolor{negativecolour}{$-$} & \textcolor{negativecolour}{ {\bf dead} damn stupid bad crazy } \\
	\cline{2-3}
	& \textcolor{positivecolour}{$+$} & \textcolor{positivecolour}{ {\bf mobile} {\bf smart} good great f***ing } \\
	\hline
	\hline
	\multirow{2}{*}{battery life} 
	& \textcolor{negativecolour}{$-$} & \textcolor{negativecolour}{ terrible poor bad horrible {\bf non-existence} } \\
	\cline{2-3}
	& \textcolor{positivecolour}{$+$} & \textcolor{positivecolour}{ good {\bf long} great {\bf 7hr} {\bf ultralong} } \\
	\hline
	\hline
	\multirow{2}{*}{game} 
	& \textcolor{negativecolour}{$-$} & \textcolor{negativecolour}{ {\bf addictive} stupid free {\bf full} {\bf addicting} } \\
	\cline{2-3}
	& \textcolor{positivecolour}{$+$} & \textcolor{positivecolour}{ great good awesome favorite {\bf cat-and-mouse} } \\
	\hline
    \hline
	\multirow{2}{*}{sausage} 
	& \textcolor{negativecolour}{$-$} & \textcolor{negativecolour}{ silly {\bf argentinian} {\bf cold} huge stupid } \\
	\cline{2-3}
	& \textcolor{positivecolour}{$+$} & \textcolor{positivecolour}{ {\bf hot grilled} good {\bf sweet} awesome } \\
	\hline
	\end{tabular}
	}
	{\small * Words in {\textbf{bold}} are more specific and can only describe certain targets.}
	\vspace{-0.15in}
\end{table}

\begin{table*}[ht!]
	\centering
	\vskip -2mm
    \caption{Aspect-based Opinion Comparison between Sony, Canon and Samsung}
    \label{tbl:exampleComparison}
	\vskip 0.15in
    \resizebox{0.88\linewidth}{!} 
    {
	\begin{tabular}{|c|c|l|l|l|}
    \hline
    \multirow{2}{*}{Brands} & \multirow{2}{*}{Sentiment} & \multicolumn{3}{c|}{Aspects / Targets' Opinions} \\
    \cline{3-5}
    & & \multicolumn{1}{c|}{Camera} & \multicolumn{1}{c|}{Phone} & \multicolumn{1}{c|}{Printer} \\
	\hline
	\multirow{4}{*}{Canon} & \multirow{2}{*}{\textcolor{negativecolour}{$-$}} 
	& \textcolor{negativecolour}{{\it camera} $\to$ expensive small bad }
	& & \textcolor{negativecolour}{{\it printer} $\to$ obscure violent digital} \\
	& & \textcolor{negativecolour}{{\it lens} $\to$ prime cheap broken }
	& & \textcolor{negativecolour}{{\it scanner} $\to$ cheap} \\
	\cline{2-5}
	& \multirow{2}{*}{\textcolor{positivecolour}{$+$}} 
	& \textcolor{positivecolour}{{\it camera} $\to$ great compact amazing }
	& & \textcolor{positivecolour}{{\it printer} $\to$ good great nice} \\
	& & \textcolor{positivecolour}{{\it pictures} $\to$ great nice creative }
	& & \textcolor{positivecolour}{{\it scanner} $\to$ great fine} \\
    \hline
	\multirow{4}{*}{Sony} & \multirow{2}{*}{\textcolor{negativecolour}{$-$}} 
	& \textcolor{negativecolour}{{\it camera} $\to$ big crappy defective }
	& \textcolor{negativecolour}{{\it phone} $\to$ worst crappy shittest}
	& \textcolor{negativecolour}{{\it printer} $\to$ stupid} \\
	& & \textcolor{negativecolour}{{\it lens} $\to$ vertical cheap wide }
	& \textcolor{negativecolour}{{\it battery life} $\to$ low }
	& \\
	\cline{2-5}
	& \multirow{2}{*}{\textcolor{positivecolour}{$+$}} 
	& \textcolor{positivecolour}{{\it photos} $\to$ great lovely amazing }
	& \textcolor{positivecolour}{{\it phone} $\to$ great smart beautiful }
	& \\
	& & \textcolor{positivecolour}{{\it camera} $\to$ good great nice }
	& \textcolor{positivecolour}{{\it reception} $\to$ perfect }
	& \\
	\hline
	\multirow{4}{*}{Samsung} & \multirow{2}{*}{\textcolor{negativecolour}{$-$}} 
	& \textcolor{negativecolour}{{\it camera} $\to$ digital free crazy }
	& \textcolor{negativecolour}{{\it phone} $\to$ stupid bad fake }
	& \textcolor{negativecolour}{{\it scanner} $\to$ worst}	 \\
	& & \textcolor{negativecolour}{{\it shots} $\to$ quick wide }
	& \textcolor{negativecolour}{{\it battery life} $\to$ solid poor terrible }
	& \\
	\cline{2-5}
	& \multirow{2}{*}{\textcolor{positivecolour}{$+$}} 
	& \textcolor{positivecolour}{{\it camera} $\to$ gorgeous great cool }
	& \textcolor{positivecolour}{{\it phone} $\to$ mobile great nice }
	& \\
	& & \textcolor{positivecolour}{{\it pics} $\to$ nice great perfect }
	& \textcolor{positivecolour}{{\it service} $\to$ good sweet friendly }
	& \\
	\hline
	\end{tabular}
	}
	\vskip -0.1in
\end{table*}

We then look at the opinion phrase distributions $\phi$'s. 
In ILDA and LDA-DP, the opinion words are generated conditioned on the latent sentiment labels, meaning that the opinion word is assumed to be independent to the target word given the sentiment;
while in TOTM, the opinion word distributions are modeled given the sentiment and the observed target word.
The advantage of TOTM over ILDA and LDA-DP in modeling the opinion words is that it allows us to analyze the opinions in a finer grained view. 
For instance, we can display a list of positive and negative opinions associated to a certain target word; an extract of this result is presented in Table~\ref{tbl:opinions}, in which we pick a few distinctive target words to show their opinion words distribution. 
As we can see from Table~\ref{tbl:opinions}, despite some opinion words can generally be applied to most target words ({\it e.g.\ }good, bad), the highlighted words are more descriptive ({\it e.g.\ }addictive, fried, grilled) and can only be applied to certain target words.
Such a result cannot be achieved by ILDA or LDA-DP.

\subsubsection{Comparing Opinions on Brands with TOTM}

We present an application of comparing opinions on entities or products using TOTM. 
%As mentioned in Subsection~\ref{subsec:preprocessing}, we group tweets containing the same tags. 
Since entities and products are frequently quoted with tags, we can compare them directly by looking at the opinions associated with each tag.
We present an extract of the opinion comparison between three brands (Canon, Sony and Samsung) in Table~\ref{tbl:exampleComparison}. 
This table shows that we can have a high level comparison of the camera product between these three brands. For the phone product, there are only comparison between Sony and Samsung, since Canon does not manufacture phones (or no tweet on such topic is found). 
Note that the entries under the aspect `{\it printer}' are lacking, we find that this is due to the low amount of opinion tweets on printers in the dataset.

\subsubsection{Extracting Contrastive Opinions on Products}

Although the above comparison is useful for providing a high level summary, it is also important to inspect the original tweets as they provide opinions in greater details.
We use TOTM to extract tweets containing people's opinions on iPhone.
In Table~\ref{tbl:contrastive}, we display an extract of contrasting tweets containing the target `iphone' with positive or negative sentiment ($r=\{-1,1\}$).

\begin{table}[t!]
	\centering
	\vskip -2.5mm
    \caption{Contrasting Opinions on iPhone}
    \label{tbl:contrastive}
	\vskip 0.15in
%    \resizebox{1.01\linewidth}{!} 
    {
	\begin{tabular}{|p{0.45\columnwidth}|p{0.45\columnwidth}|}
	\hline
	\multicolumn{1}{|c|}{Positive} & \multicolumn{1}{c|}{Negative} \\
	\hline
	\scriptsize RT @user : the iPhone is so awesome!!! Emailing, texting, surfing the sametime! --- Can do all tgat while talkin on the phone?...
	& \scriptsize @user awww thx! I can't send an email right now bc my iPhone is stupid with sending emails. Lol but I can tweet or dm u? \\
	\hline
	\scriptsize Ahhh! Tweeting on my gorgeous iPhone! I missed you! hehe am on my way home, put the kettle on will you pls : )
	& \scriptsize It would appear that the iPhone, due to construction, is weak at holding signal. Combine that with a bullshit 3G network in Denver. \\
	\hline
	\scriptsize Thanks @user for the link to iPhone vs Blackberry debate. I got the iPhone \& it's just magic! So intuitive!
	& \scriptsize @user @user Ah, well there you go. The iPhone is dead, long live Android! ;) \\ 
	\hline
	\scriptsize Finally my fave lover @user has Twitter \& will be using it all the time with her cool new iPhone :)
	& \scriptsize @user Finally eh? :D I think iphone is so ugly x.x \\
	\hline
	\end{tabular}
	}
	\vspace{-0.15in}
\end{table}

%\subsection{Diagnostic}
%-diagnostic plots -> likelihood convergence - cold start vs warm start?
%-hyperparameters convergence and mixing.

%\subsection{Further applications?}

%\vspace{-1mm}
\section{Conclusion}
\label{sec:conclusion}

In this paper, we study the use of LDA-based models for opinion analysis on tweets queried with electronic product terms. 
This is motivated by the fact that Twitter is a popular platform for opinions and tweets are publicly available.
%Also, we speculate that tweets are less likely to be targeted by companies to spread fake reviews. 
Unlike reviews, tweets do not contain scores or ratings, they are more informal and usually accompanied by emoticons and strong sentiment words. 
Taking advantage of the informal nature of tweets, we designed a topic model named Twitter Opinion Topic Model (TOTM) for opinion analysis. 
TOTM is shown to greatly improve opinion prediction with the direct target-opinion modeling.
In incorporating a sentiment lexicon into topic models, we proposed a new formulation for the topic 
model priors, which learns and updates given data. Our innovative formulation is shown to improve 
sentiment analysis significantly.

Our qualitative analysis demonstrates that opinion mining on tweets provide useful opinions on electronic products.
Note that although we can obtain a large quantity of product opinions on tweets, the opinions are usually much noisier than reviews.
For instance, opinions can be incidental ({\it e.g.}\ the author was just frustrated with the product that time), since it is easy and effortless to produce a tweet.
%As a note of caution, the opinions extracted from Twitter do not represent the public opinions, hence one needs to be cautious when interpreting the result. 
As with the reviews, the opinions on tweets may not always be true.
Some tweets are laden with sarcasm, making them difficult to interpret, while some others are spam containing no useful information. 

We emphasize the importance of preprocessing steps. For instance, word normalization allows 
misspellings and abbreviations to be captured for target-opinion analysis; tweet aggregation 
improves aspect clustering and lets us compare different products or brands.
For practical applications, filtering sarcastic tweets and spam is also important. 
In this paper, we have attempted to filter spam by removing tweets containing URLs. 
We acknowledge that although there is existing work on removing sarcastic tweets and 
spam~\cite{tsur2010icwsm, mccord2011spam}, we did not incorporate them due to the lack of publicly 
available software.
As future work, we are interested in utilizing other word lexicons such as synonym and antonym 
lexicons into an LDA-based model for sentiment analysis.

%1. Not everybody is using Twitter.
%Social media is not a representative
%and unbiased sample of the voting population. Some strata are under-
%represented while others are over-represented in Twitter. Demographic
%bias should be acknowledged and predictions corrected on its basis.
%2. Not every twitterer is tweeting about politics. A minority of users are
%responsible for most of the political chatter and, thus, their opinions will
%drive what can be predicted from social media. This self-selection bias is
%still an open problem.
%3. Just because it is on Twitter does not mean it is true. A substantial
%amount of data is not trustworthy and, thus, it should be discarded. There
%is a growing body of work in this regard but it is not being widely applied
%when trying to predict elections.
%4. Naïveté is not bliss.
%Simplistic sentiment analysis methods should be
%avoided one and for all. Political discourse is plagued with humor, dou-
%ble entendres, and sarcasm; this makes determining political preference of
%users hard and inferring voting intention even harder. (cite Gayo's paper)
%
%Sarcasm

\section{Acknowledgments}
%We would like to thank the anonymous reviewers for their helpful feedback and comments.
NICTA is funded by the Australian Government through the Department of
Communications and the Australian Research Council through the ICT
Centre of Excellence Program.
We also like to thank Scott Sanner, Shamin Kinathil, Rishi Dua 
and the anonymous reviewers for their feedback and comments.


%\bibliographystyle{abbrv}
%\bibliography{bibliography/biblio} 

% ACM needs 'a single self-contained file'!

\end{document}